%% file: paper2450.tex
\documentclass[runningheads]{llncs}
\usepackage[T1]{fontenc}
\usepackage{graphicx}
\usepackage{color}
\usepackage{hyperref}

\usepackage{times}
\usepackage{epsfig}
\usepackage{amsmath}
\usepackage{amssymb}
\usepackage{bm}
\usepackage{booktabs}
\usepackage{tablefootnote}
\usepackage{color}

\usepackage{tikz}
\usepackage{pifont}

\newcommand{\resultsection}[1]{\subsubsection{#1}}
\newcommand{\datasection}[1]{\textbf{#1}}

\begin{document}
\title{Prompt-MIL: Boosting Multi-Instance Learning Schemes via Task-specific Prompt Tuning}

\titlerunning{Prompt-MIL: Boosting MIL Schemes via Prompt Tuning}

\author{Jingwei Zhang\inst{1} \and
Saarthak Kapse\inst{1} \and Ke Ma \inst{2} \and Prateek Prasanna\inst{1} \and
Joel Saltz\inst{1} \and Maria Vakalopoulou\inst{3} 
\and Dimitris Samaras\inst{1}}
\authorrunning{J. Zhang et al.}

\institute{
    Stony Brook University, USA \and Snap Inc., USA
        \and
        CentraleSupélec, University of Paris-Saclay, France\\
    \email{\email{\{jingwezhang, kemma, samaras\}@cs.stonybrook.edu}} \\
    \email{\{saarthak.kapse, prateek.prasanna\}@stonybrook.edu} \\
    \email{Joel.Saltz@stonybrookmedicine.edu   maria.vakalopoulou@centralesupelec.fr}
}



%
\maketitle              
\begin{abstract}
Whole slide image (WSI) classification is a critical task in computational pathology, requiring the processing of gigapixel-sized images, which is challenging for current deep-learning methods. 
Current state of the art methods are based on multi-instance learning schemes (MIL), which usually rely on pretrained features to represent the instances. 
Due to the lack of task-specific annotated data, these features are either obtained from well-established backbones on natural images, or, more recently from self-supervised models pretrained on histopathology. 
However, both approaches yield task-agnostic features, resulting in performance loss compared to the appropriate task-related supervision, if available.
In this paper, we show that when task-specific annotations are limited, we can inject such supervision into downstream task training, to reduce the gap between fully task-tuned and task agnostic features. 
We propose Prompt-MIL, an MIL framework that integrates prompts into WSI classification. 
Prompt-MIL adopts a prompt tuning mechanism, where only a small fraction of parameters calibrates the pretrained features to encode task-specific information, rather than the conventional full fine-tuning approaches.
Extensive experiments on three WSI datasets, TCGA-BRCA, TCGA-CRC, and BRIGHT, demonstrate the superiority of Prompt-MIL over conventional MIL methods, achieving a relative improvement of 1.49\%-4.03\% in accuracy and 0.25\%-8.97\% in AUROC while using fewer than 0.3\% additional parameters. 
Compared to conventional full fine-tuning approaches, we fine-tune less than 1.3\% of the parameters, yet achieve a relative improvement of 1.29\%-13.61\% in accuracy and 3.22\%-27.18\% in AUROC and reduce GPU memory consumption by 38\%-45\% while training 21\%-27\% faster.

\end{abstract}

\input{1_intro}
\input{2_method}

\input{3_exp}

\input{4_sum}

\subsubsection{Acknowledgements} This work was partially supported by the ANR Hagnodice ANR-21-CE45-0007, the NSF IIS-2212046, the NSF IIS-2123920, the NIH 1R21CA258493-01A1, the NCI UH3CA225021 and Stony Brook University Provost Funds. The content is solely the responsibility of the authors and does not necessarily represent the official views of the National Institutes of Health.

%
%
%
\bibliographystyle{splncs04}
\bibliography{bibliography}
\end{document}


\title{Supplementary Material of \\ Prompt-MIL: Boosting Multi-Instance Learning Schemes via Task-specific Prompt Tuning}

\titlerunning{}

\author{}
\authorrunning{}
\institute{}
\maketitle              

\begin{table}[h]
\caption{Comparison of accuracy and AUROC between the conventional MIL model and fine-tuning the last Transfromer encoder ($L_{12}$ in ViT-T), using various MIL schemes. ``Num. of Parameters'' represents the number of optimized parameters.}
\label{table:sup:mil}
\begin{center}
\setlength{\tabcolsep}{0.9mm}{

\begin{tabular}{c | l | c c c c |c}
\toprule
\multicolumn{1}{c|}{MIL scheme} 
&\multicolumn{1}{c|}{Method} 
        & \multicolumn{2}{c}{TCGA-CRC}
            & \multicolumn{2}{c}{BRIGHT}
    & \multicolumn{1}{|c}{Num. of}
\\
&\multicolumn{1}{c|}{} 
        & \multicolumn{1}{c}{Accuracy} & \multicolumn{1}{c}{AUROC}
            & \multicolumn{1}{c}{Accuracy} & \multicolumn{1}{c}{AUROC}
    & \multicolumn{1}{|c}{Parameters} 
\\
\midrule
&Conventional MIL 
        & $73.02$         & $69.24$
            & $62.08$         & $80.96$
    & 64k
\\
DSMIL~\cite{li2021dual_dsmil} & Fine-tuning $L_{12}$
        & 69.81  & 70.72
            & 61.25 & 80.10
    & 509k
\\ 
&Prompt-MIL (ours) 
        & $\bm{75.47}$  & $\bm{75.45}$ 
            & $\bm{64.58}$  & $\bm{81.31}$ 
            
    & 64k+192
\\
\midrule
&Conventional MIL 
        & $74.10$         & $68.56$
            & $61.25$         & $\bm{80.35}$
    & 25k
\\
ABMIL~\cite{ilse2018attention_abmil} & Fine-tuning $L_{12}$
        & 70.37  &  $\bm{70.78}$
            & $\bm{62.50}$ & 78.92
    & 470k
\\ 
&Prompt-MIL (ours) 
        & $\bm{75.87}$  & $\bm{70.10}$ 
            & $\bm{62.50}$  & $79.30$ 
            
    & 25k+192
\\
\midrule
&Conventional MIL 
        & $75.87$         & $77.50$
            & $62.08$         & $82.97$
    & 59k
\\
CLAM~\cite{lu2021data_clam} &Fine-tuning $L_{12}$
        &  74.07 & $71.40$ 
            & 63.33  & 81.32
    & 504k
\\ 
&Prompt-MIL (ours) 
        & $\bm{76.19}$  & $\bm{80.84}$ 
            & $\bm{64.17}$  & $\bm{84.31}$ 
            
    & 59k+192
\\
            
\bottomrule
\end{tabular}
}
\end{center}
\end{table}

\begin{table}[hb]
\caption{Comparison of accuracy and AUROC on TCGA-BRCA and BRIGHT between the baseline ViT-small model and the same model fine-tuned by our Prompt-MIL. The baseline ViT-small model is a pathological foundation model, which was pretrained using DINO and the entire TCGA dataset~\cite{chen2022scaling_hipt}. It is a different model from the one in the main paper.}
\label{table:result:universal_models}
\begin{center}
\setlength{\tabcolsep}{1.6mm}{
\begin{tabular}{l c c  c c }
\toprule
\multicolumn{1}{c}{Dataset} 
    & \multicolumn{2}{c}{TCGA-BRCA} 
            & \multicolumn{2}{c}{BRIGHT} \\
\multicolumn{1}{c}{Metric} 
& \multicolumn{1}{c}{Accuracy} & \multicolumn{1}{c}{AUROC} 
& \multicolumn{1}{c}{Accuracy} & \multicolumn{1}{c}{AUROC}  \\
\midrule
ViT-small in ~\cite{chen2022scaling_hipt}
    & $88.49$          & $90.69$  
            & $56.25$  & $73.69$ \\
ViT-small w/ Prompt-MIL
    & $\bm{92.00} $ & $\bm{95.65}$
            & $\bm{60.00}$  & $\bm{75.79}$ \\
\bottomrule
\end{tabular}
}
\end{center}
\end{table}

\begin{figure}[t]
\begin{center}
\includegraphics[width=\linewidth]{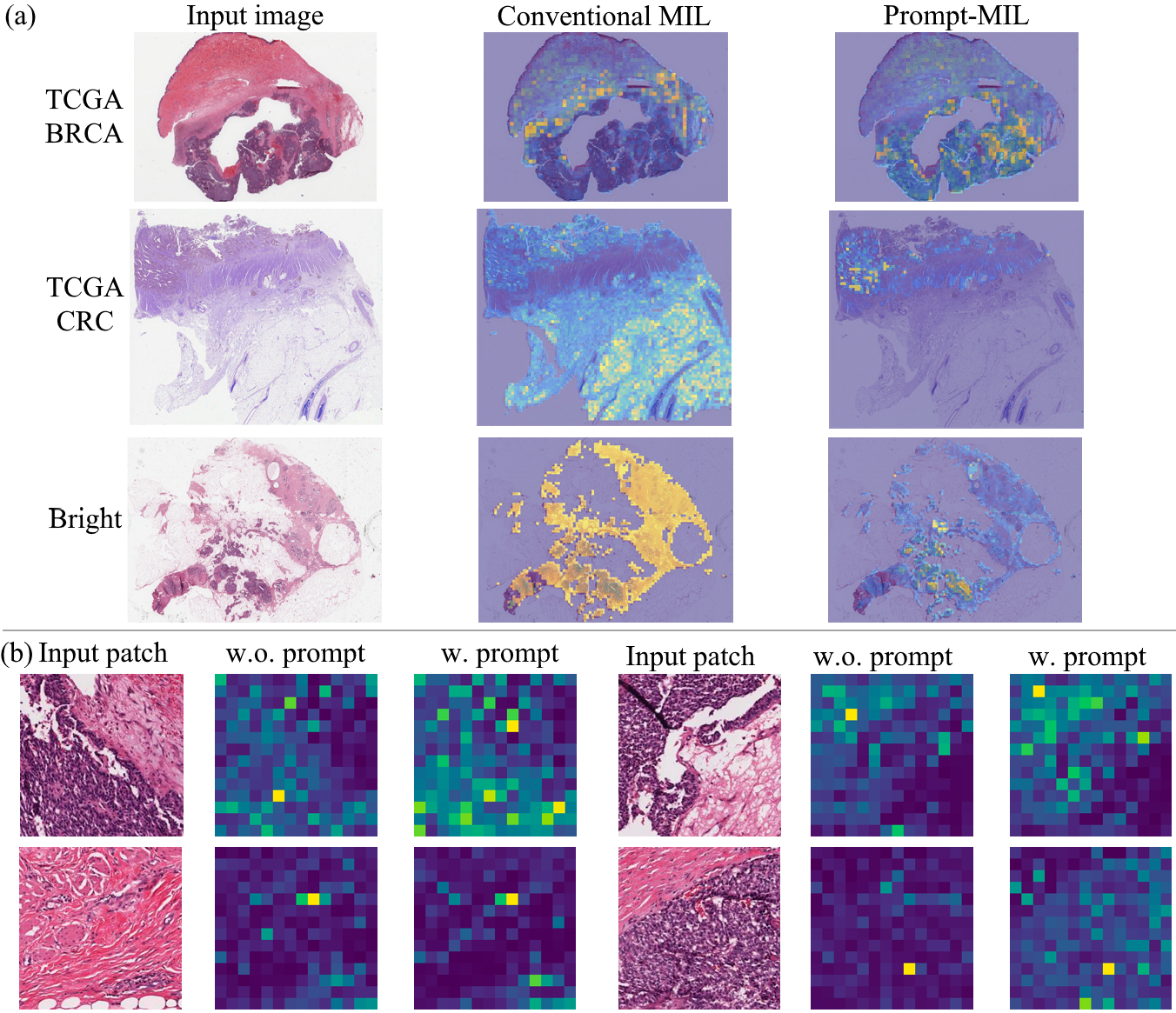}
\end{center}
   \caption{(a) The attention visualization of the classifier $G(\cdot)$. Compared to that in the conventional MIL, the attention map of our Prompt-MIL focused more on the tumor regions which are critical for the cancer classification task. (b) The attention visualization of the last multi-head self-attention layer in the feature model $F(\cdot)$. The prompt token guided the attention to cover more on the tumor regions.}
\label{fig:vis}
\end{figure}


\vspace{0.8\baselineskip}
\bibliographystyle{splncs04}
\bibliography{supplement.bib}

%% file: 1_intro.tex
\section{Introduction}

Whole slide image (WSI) classification is a critical task in computational pathology enabling disease diagnosis and subtyping using automatic tools. 
Owing to the paucity of patch-level annotations, multiple instance learning (MIL) ~\cite{hou2016patch,lu2021data_clam,shao2021transmil} techniques have become a staple in WSI classification. 
Under an MIL scheme, WSIs are divided into tissue patches or instances, and a feature extractor is used to generate features for each instance. 
These features are then aggregated using different pooling or attention-based operators to provide a WSI-level prediction.
ImageNet pretrained networks have been widely used as MIL feature extractors. 
More recently, self-supervised learning (SSL), using a large amount of unlabeled histopathology data, has become quite popular for WSI classification~\cite{li2021dual_dsmil,chen2022scaling_hipt} as it outperforms ImageNet feature encoders.

Most existing MIL methods do not fine-tune their feature extractor together with their classification task; this stems from the requirement for far larger GPU memory than is available currently due to the gigapixel nature of WSIs, e.g. training a WSI at 10x magnification may require more than 300 Gb of GPU memory.
Recently, researchers have started to explore optimization methods to enable end-to-end training of the entire network and entire WSI within GPU memory~\cite{takahama2019multi_retaining,pinckaers2020streaming,zhang2022gigapixel_local_learning}.
These methods show better performance compared to conventional MIL; they suffer, however, from two limitations. 
First, they are ImageNet-pretrained and do not leverage the powerful learning capabilities of histology-trained SSL models. 
Second, these are mostly limited to convolutional architectures rather than more effective attention-based architectures such as vision transformers~\cite{vit}.

\textbf{Motivation:} To improve WSI-level analysis, we explore end-to-end training of the entire network using SSL pretrained ViTs. To achieve this, we use the patch batching and gradient retaining techniques in~\cite{takahama2019multi_retaining}. 
However, we find that conventional fine-tuning approaches, where the entire network is fine-tuned, achieve low performance.
For example, on the BRIGHT dataset\cite{bright}, the accuracy drops more than 5\% compared to the conventional MIL approaches.
The poor performance is probably caused by the large network over-fitted to the limited downstream training data, leading to suboptimal feature representation. 
Indeed, especially for weakly supervised WSI classification, where annotated data for downstream tasks is significantly less compared to natural image datasets, conventional fine-tuning schemes can prove to be quite challenging.

To address the subpar performance of SSL-pretrained vision transformers, 
we utilize the prompt tuning techniques.
Initially proposed in natural language processing, a prompt is a trainable or a pre-defined natural language statement that is provided as additional input to a transformer to guide the neural network towards learning a specific task or objective~\cite{brown2020language_prompt_gpt,lester2021power_prompt_org}. 
Using prompt tuning we \textit{fine-tune only the prompt and downstream network without re-training the large backbone} (e.g. GPT-3 with 17B parameters). 
This approach is parameter efficient~\cite{lester2021power_prompt_org,liu2022p_p_tuning} and has been shown to better inject task-specific information and reduce the overfitting in downstream tasks, particularly in limited data scenarios~\cite{schucher2022power_prompt_lwo_res,gu2022ppt_prompt_few_shot}. 
Recently, prompts have also been adopted in computer vision and demonstrated superior performance compared to conventional fine-tuning methods~\cite{jia2022visual_prompt_vpt}.
Prompt tuning performs well even when only limited labeled data is available for training, making it particularly attractive in computational pathology. 
The process of prompt tuning thus involves providing a form of limited guidance during the training of downstream tasks, with the goal of minimizing the discrepancy between feature representations that are fully tuned to the task and those that are not task-specific. 


In this paper, we propose a novel framework, Prompt-MIL, which uses prompts for WSI-level classification tasks within an MIL paradigm. Our contributions are: 
\begin{itemize}
    \item \textbf{Fine-tuning:} Unlike existing works in histopathology image analysis, Prompt-MIL is fine-tuned using prompts rather than conventional full fine-tuning methods.
    \item \textbf{Task-specific representation learning:} Our framework employs an SSL pretrained ViT feature extractor with a trainable prompt that calibrates the representations making them task-specific. By doing so, only the prompt parameters together with the classifier, are optimized. This avoids potential overfitting while still injecting task-specific knowledge into the learned representations.
\end{itemize}
Extensive experiments on three public WSI datasets, TCGA-BRCA, TCGA-CRC, and BRIGHT demonstrate the superiority of Prompt-MIL over conventional MIL methods, achieving a relative improvement of 1.49\%-4.03\% in accuracy and 0.25\%-8.97\% in AUROC by using only less than 0.3\% additional parameters. 
Compared to the conventional full fine-tuning approach, we fine-tune less than 1.3\% of the parameters, yet achieve a relative improvement of 1.29\%-13.61\% in accuracy and 3.22\%-27.18\% in AUROC. 
Moreover, compared to the full fine-tuning approach, our method reduces GPU memory consumption by 38\%-45\% and trains 21\%-27\% faster. 
To the best of our knowledge, this is the first work where prompts are explored for WSI classification. 
While our method is quite simple, it is versatile as it is agnostic to the MIL scheme and can be easily applied to different MIL methods.
Our code is available at \href{https://github.com/cvlab-stonybrook/PromptMIL}{https://github.com/cvlab-stonybrook/PromptMIL}.

%% file: 2_method.tex
\section{Method}

\begin{figure}[t]
\begin{center}
\includegraphics[width=\linewidth]{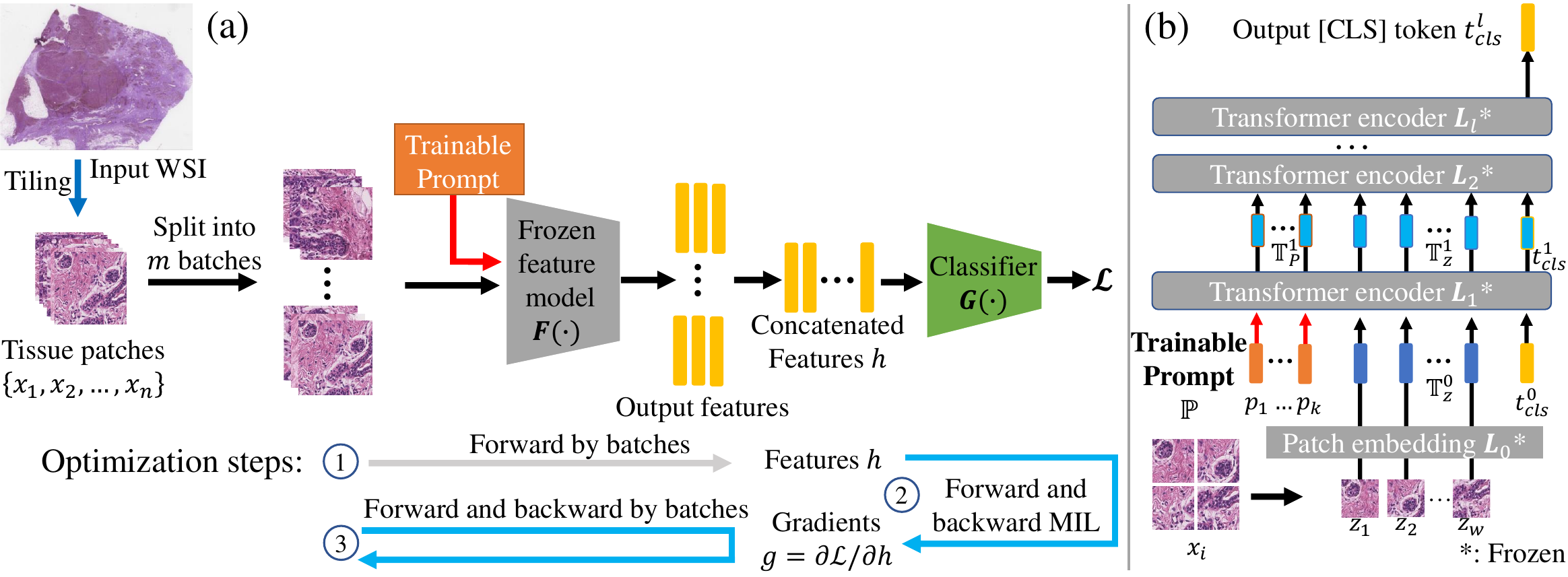}
\end{center}
   \caption{Overview of the proposed method.
   (a) Overall structure of our training pipeline. 
   Tissue patches tiled from the input WSI are grouped into separate batches, which are fed into a frozen feature model $F(\cdot)$ to compute their respective features. 
   The features are subsequently concatenated into the feature $h$ and a classifier $G(\cdot)$ applies an MIL scheme on $h$ to predict the label and calculate the loss $\mathcal{L}$. 
   (b) Structure of the feature model $F(\cdot)$ with the additional prompt. An input image $x_i$ is cropped into $w$ small patches $z_1, \dots, z_w$. $k$ trainable prompt tokens, together with the embedding of small patches and a class token $t_{cls}^0$, are fed into $l$ layers of Transformer encoders. The output feature corresponding to $x_i$ is the last class token $t_{cls}^{l}$. The feature model $F(\cdot)$ is frozen and only the prompt is trainable.
   }
\label{fig:framework}
\end{figure}

Our Prompt-MIL framework consists of three components: a frozen feature model to extract features of tissue patches, a classifier that performs an MIL scheme of feature aggregation and classification of the WSIs, and a trainable prompt. 
Given a WSI and its label $y$, the image is tiled into $n$ tissue patches/instances $\{x_1, x_2, \dots, x_n\}$ at a predefined magnification.
As shown in Fig. \ref{fig:framework}, the feature model $F(\cdot)$ computes $n$ feature representations from the corresponding $n$ patches:
\begin{equation}
    \begin{aligned}
    h &= [h_1, h_2, \dots, h_n] \\
    &= [F(x_1, \mathbb{P}), F(x_2, \mathbb{P}),\dots, F(x_n, \mathbb{P})],
    \end{aligned}
\end{equation}
where $h_i$ denotes the feature of the $i^{th}$ patch, $h$ is the concatenation of all $h_i$, and $\mathbb{P} = \{p_i, i = 1,2,\dots, k\}$ is the trainable prompt consisting of $k$ trainable tokens. 
The classifier $G(\cdot)$ applies an MIL scheme to predict the label $\hat{y}$ and calculate the loss $\mathcal{L}$ as:
\begin{align}
    \mathcal{L} &= \mathcal{L}_{cls}(\hat{y}, y) = \mathcal{L}_{cls}(G(h), y),
\end{align}
where the $\mathcal{L}_{cls}$ is a classification loss.

\subsection{Visual prompt tuning}
The visual prompt tuning is the key component of our framework.
As shown in Fig.~\ref{fig:framework}(b), our feature model $F(\cdot)$ is a ViT based architecture.
It consists of a patch embedding layer $L_0$ and $l$ sequential encoding layers \{$L_1$, $L_2$, \dots, $L_l$\}.
The ViT first divides an input image $x_i$ into $w$ smaller patches $[z_1, z_2, \dots,z_w]$ and embeds them into $w$ tokens:
\begin{align}
    \mathbb{T}_z^0 
        &= L_0([z_1, z_2, \dots,z_w]) = \{t_1^0, t_2^0, \dots, t_w^0\},
\end{align}
where $t_i^0$ is the embedding token of $z_i$ and $\mathbb{T}_z^0$ is the collection of such tokens. 
These tokens $\mathbb{T}_z^0$ are concatenated with a class token $t_{cls}^0$ and a prompt $\mathbb{P}$:
The class token is used to aggregate information from all other tokens.
The prompt consists of $k$ trainable tokens $\mathbb{P} = \{p_i| i = 1,2,\dots, k\}$.
The concatenation is fed into $l$ layers of the Transformer encoders:
\begin{align}
    [\mathbb{T}_z^1, \mathbb{T}_P^1, t_{cls}^1] &= L_1([\mathbb{T}_z^0, \mathbb{P}, t_{cls}^0]) 
    \\
    [\mathbb{T}_z^i, \mathbb{T}_P^i, t_{cls}^i] &= L_i([\mathbb{T}_z^{i-1}, \mathbb{T}_P^{i-1}, t_{cls}^{i-1}]), i = 2, 3, \dots, l
    \\
    \mathbb{T}_P^i &= \{p_j^i| j = 1,2,\dots, k\},
\end{align}
where $p_j^i$ is the $j^{th}$ output prompt token of the $i^{th}$ Transformer encoder and $\mathbb{T}_P^i$ is the collection of all $k$ such output prompt tokens, which are not trainable.
The output feature of $x_i$ is defined as the last class token: $h_i = t_{cls}^{l}$.

\subsection{Optimization}
Our overall loss function is defined as
\begin{equation}
    \begin{aligned}
    \mathcal{L} &= \mathcal{L}_{cls}(G(H), y) \\
    &= \mathcal{L}_{cls}(G([F(x_1, \mathbb{P}), F(x_2, \mathbb{P}),\dots, F(x_n, \mathbb{P})]), y),
    \end{aligned}
\end{equation}
where only the parameters of the $G(\cdot)$ and the prompt $\mathbb{P}$ are optimized, while the feature extractor model $F(\cdot)$ is frozen.

Training the entire pipeline in an end-to-end fashion on gigapixel images is infeasible using the current hardware. 
To address this issue, we utilize the patch batching and gradient retaining techniques from~\cite{takahama2019multi_retaining}.
As shown in Fig.\ref{fig:framework}(a), to reduce the GPU memory consumption, the $n$ tissue patches $\{x_1, x_2,\dots, x_n\}$ are grouped into $m$ batches. 
The first step (step\ding{172} in the figure) of our optimization is to sequentially feed $m$ batches of tissue patches forward to the feature model to compute its respective features which are subsequently concatenated into the $h$ matrix.
In this step, we just conduct a forward pass like the inference stage, without storing the memory-intensive computational graph for back-propagation.

In the second step (step\ding{173}), we feed $h$ into the classifier $G(\cdot)$ to calculate the loss $\mathcal{L}$ and update the parameters of $G(\cdot)$ by back-propagate the loss. 
The back-propagated gradients $g=\partial{\mathcal{L}}/\partial{h}$ on $h$ are retained for the next step.

Finally (step\ding{174}), 
we feed the input batches into the feature model $F(\cdot)$ again and use the output $h$ and the retained gradients $g$ from the last step to update the trainable prompt tokens.
In particular, the gradients on the $j^{th}$ prompt token $p_j$ are calculated as:
\begin{equation}
    \begin{aligned}
    \frac{\partial \mathcal{L}}{\partial p_j} 
    &= \frac{\partial \mathcal{L}}{\partial h} \frac{\partial h}{\partial p_j} \\
    &= \sum_i \frac{\partial \mathcal{L}}{\partial h_i} \frac{\partial h_i}{\partial p_j}  
    = \sum_i g_i \frac{\partial h_i}{\partial p_j},
    \end{aligned}
\end{equation}
where $g_i$ is the gradient calculated with respect to $h_i$.

To sum up, in each step, we only update either $F$ or $G$ given the current batch, which avoid storing the gradients of the whole framework for all the input patches. 
This patch batching and gradient retaining techniques make the end-to-end training feasible.

In this study, we use DSMIL~\cite{li2021dual_dsmil} as the classifier and binary cross entropy as the classification loss $\mathcal{L}_{cls}$ when the task is a tumor sub-type classification or cross entropy otherwise. 

%% file: 3_exp.tex
\section{Experiments and Discussion}
\subsection{Datasets}
We assessed Prompt-MIL using three histopathological WSI datasets: TCGA-BRCA \cite{tcga_brca}, TCGA-CRC~\cite{cancer2012comprehensive}, and BRIGHT~\cite{bright}.
These datasets were utilized for both the self-supervised feature extractor pretraining and the end-to-end fine-tuning (with or without prompts), including the MIL component. 
Note that the testing data were not used in the SSL pretraining. \datasection{TCGA-BRCA} contains 1034 diagnostic digital slides of two breast cancer subtypes: invasive ductal carcinoma (IDC) and invasive lobular carcinoma (ILC). 
We used the same training, validation, and test split as that in the first fold cross validation in~\cite{chen2022scaling_hipt}. 
The cropped patches (790K training, 90K test) were extracted at 5$\times$  magnification. 
\datasection{TCGA-CRC} contains 430 diagnostic digital slides of colorectal cancer for a binary classification task: chromosomal instability (CIN) or genome stable (GS). 
Following the common 4-fold data split~\cite{bilal2021development,liu2018comparative}, we used the first three folds for training (236 GS, 89 CIN), and the fourth for testing (77 GS, 28 CIN). 
We further split 20\% (65 slides) training data as a validation set. The cropped patches (1.07M training, 370K test) were extracted at 10$\times$  magnification. 
\datasection{BRIGHT} contains 503 diagnostic slides of breast tissues. 
We used the official training (423 WSIs) and test (80 WSIs) splits. 
The task involves classifying non-cancerous (196 training, 25 test) vs. pre-cancerous (66 training, 23 test) vs. cancerous (161 training, 32 test). 
We further used 20\% (85 slides) training slides for validation. 
The cropped patches (1.24M training, 195K test) were extracted at 10$\times$ magnification. 

\begin{table}[t]
\caption{Comparison of accuracy and AUROC on three datasets. Reported metrics (in $\%$age) are the average across 3 runs. "Num. of Parameters" represents the number of optimized parameters}
\label{table:result:accuracy}
\begin{center}
\setlength{\tabcolsep}{0.9mm}{

\begin{tabular}{l |c c c c c c |c}
\toprule
\multicolumn{1}{c|}{Dataset} 
    & \multicolumn{2}{c}{TCGA-BRCA} 
        & \multicolumn{2}{c}{TCGA-CRC}
            & \multicolumn{2}{c}{BRIGHT}
    & \multicolumn{1}{|c}{Num. of}
\\
\multicolumn{1}{c|}{Metric} 
    & \multicolumn{1}{c}{Accuracy} & \multicolumn{1}{c}{AUROC} 
        & \multicolumn{1}{c}{Accuracy} & \multicolumn{1}{c}{AUROC}
            & \multicolumn{1}{c}{Accuracy} & \multicolumn{1}{c}{AUROC}
    & \multicolumn{1}{|c}{Parameters} 
\\
\midrule
Conventional MIL 
    & $92.10$          & $96.65$       
        & $73.02$         & $69.24$
            & $62.08$         & $80.96$
    & 70k
\\
Full fine-tuning 
    & $88.14$ & $93.78$
        & $74.53$ & $56.63$
            & $56.13$ & $75.87$
    & 5.6M
\\ 
Prompt-MIL (ours) 
    & $\bm{93.47}$ & $\bm{96.89}$
        & $\bm{75.47}$  & $\bm{75.45}$ 
            & $\bm{64.58}$  & $\bm{81.31}$ 
            
    & 70k+192
\\
\bottomrule
\end{tabular}
}
\end{center}
\end{table}

\subsection{Implementation Details}
We cropped non-overlapping 224 $\times$ 224 sized patches in all our experiments and used ViT-Tiny (ViT-T/16)~\cite{vit} for feature extraction.
For SSL pretraining, we leveraged the DINO framework~\cite{dino} with the default hyperparameters, but adjusted the batch size to 256 and employed the global average pooling for token aggregation. 
We pretrained separate ViT models on the TCGA-CRC datasets for 50 epochs, on the BRIGHT dataset for 50 epochs, and on the BRCA dataset for 30 epochs. 
For TCGA-BRCA, we used the AdamW~\cite{loshchilov2017adamW} optimizer with a learning rate of $1e-4$, $1e-2$ weight decay, and trained for 40 epochs.
For TCGA-CRC, we also used the AdamW optimizer with a learning rate of $5e-4$ and trained for 40 epochs.
For Bright, we used the Adam~\cite{adam} optimizer with a learning rate of $1e-4$, $5e-2$ weight decay and trained for 40 epochs.
We applied a cosine annealing learning rate decay policy in all our experiments.
For the MIL baselines, we employed the same hyperparameters as above.
For all full fine-tuning experiments, we used the learning rate in the corresponding prompt experiment as the base learning rate. For parameters in the feature model $F(\cdot)$, which are SSL pretrained, we use 1/10 of the base learning rate. For parameters in the Classifier $G(\cdot)$, which are randomly initialized, we use the base learning rate. We train the full tuning model for 10 more epochs than our prompt training to allow full convergence. This training strategy is optimized using the validation datasets.
All model implementations were in PyTorch~\cite{paszke2019pytorch} on a NVIDIA Tesla V100 or a Nvidia Quadro RTX 8000.

\subsection{Results}
We chose overall accuracy and Area Under Receiver Operating Characteristic curve (AUROC) as the evaluation metrics.

\resultsection{Evaluation of prompt tuning performance:} \label{sec:result_prompt}
We compared the proposed Prompt-MIL with two baselines: 1) a conventional MIL model with a frozen feature extractor~\cite{li2021dual_dsmil}, 2) fine-tuning all parameters in the feature model (full fine-tuning).
Table~\ref{table:result:accuracy} highlights that our Prompt-MIL consistently outperformed both.
Compared to the conventional MIL method, Prompt-MIL added negligible parameters (192, less than 0.3\% of the total parameters), 
achieving a relative improvement of 1.49\% in accuracy and 0.25\% in AUROC on TCGA-BRCA, 3.36\% in accuracy and 8.97\% in AUROC on TCGA-CRC, and 4.03\% in accuracy and 0.43\% in AUROC on BRIGHT.
The observed improvement can be attributed to a more optimal alignment between the feature representation learned during the SSL pretraining and the downstream task, i.e., the prompt explicitly calibrated the features toward the downstream task.

The computationally intensive full fine-tuning method under-performed conventional MIL and Prompt-MIL. 
Compared to the full fine-tuning method, our method achieved a relative improvement of 1.29\% to 13.61\% in accuracy and 3.22\% to 27.18\% in AUROC on the three datasets.
Due to the relatively small amount of slide-level labels (few hundred to a few thousands) fully fine tuning 5M parameters in the feature model might suffer from overfitting. 
In contrast, our method contained less than 1.3\% of parameters compared to full fine-tuning, leading to robust training.

\begin{table}[t]
\caption{Comparison of GPU memory consumption and training speed per slide benchmarked on the BRIGHT dataset between the full fine-tuning and our prompt tuning on four slides with different sizes. Our prompt method requires far less memory and is significantly faster.}  
\begin{center}
\setlength{\tabcolsep}{1.6mm}{
\begin{tabular}{clcccc}
\toprule
\multicolumn{2}{c}{WSI size}  
    & $44k\times21k$ & $26k\times21k$ 
        & $22k\times17k$ & $11k\times16k$ \\
\multicolumn{2}{c}{\#Tissue patches} 
    & 9212               & 4765               
        & 2307               & 1108               \\ 
\midrule
GPU        & Full fine-tuning 
    & 21.81G             & 18.22G             
        & 16.37G             & 12.71G             \\
Mem.     & Prompt (ours)    
    & $\bm{12.04}$G             & $\bm{10.66}$G             
        & $\bm{10.00}$G              & $\bm{7.90}$G              \\ 
\cmidrule{2-6}
        & Reduction percentage 
    & 44.79\%  & 41.50\% & 38.92\% & 37.84\% \\
\midrule
Time      & Full fine-tuning 
    & 17.73s             & 8.92s              
        & 4.37s              & 2.15s              \\
per slide  & Prompt (ours)    
    & $\bm{13.92}$s             & $\bm{7.09}$s              
        & $\bm{3.35}$s              & $\bm{1.56}$s              \\
\cmidrule{2-6}
& Reduction percentage 
    & 21.49\%  & 20.51\% & 23.32\% & 27.27\% \\
\bottomrule
\end{tabular}
}
\end{center}
\label{table:result:gpu}
\end{table}

\begin{table}[b]
\caption{Comparison of accuracy and AUROC on three datasets for a pathological foundation model.
}
\label{table:result:universal_models}
\begin{center}

\begin{tabular}{l c c c c }
\toprule
\multicolumn{1}{c}{Dataset} & \multicolumn{2}{c}{TCGA-BRCA} & \multicolumn{2}{c}{BRIGHT} \\
\multicolumn{1}{c}{Metric} & \multicolumn{1}{c}{Accuracy} & \multicolumn{1}{c}{AUROC} & \multicolumn{1}{c}{Accuracy} & \multicolumn{1}{c}{AUROC}  \\
\midrule
ViT-small~\cite{wang2022transformer}
    & $91.75$          & $97.03$       
        & $54.17$  & $76.76$ \\
ViT-small w/ Prompt-MIL
    & $\bm{92.78} $ & $\bm{97.53}$
        & $\bm{57.50}$  & $\bm{78.29}$ \\
\bottomrule
\end{tabular}
\end{center}
\end{table}

\resultsection{Evaluation of time and GPU memory efficiency:} Prompt-MIL is an efficient method requiring less GPU memory to train and running much faster than full fine-tuning methods. 
We evaluated the training speed and memory consumption of our method and compared to the full fine-tuning baseline on four different sized WSIs in the BRIGHT dataset.
As shown in Table~\ref{table:result:gpu}, our method consumed around 38\% to 45\% less GPU memory compared to full fine-tuning and was 21\% to 27\% faster. 
As we scaled up the WSI size (i.e. WSIs with more number of patches), the memory cost difference between Prompt-MIL and full fine-tuning further widened. 

\resultsection{Evaluation on the pathological foundation models:} 
We demonstrated our Prompt-MIL also had a better performance when used with the pathological foundation model.
Foundational models refer to those trained on large-scale pathology datasets (e.g. the entire TCGA Pan-cancer dataset~\cite{weinstein2013cancer}). 
We utilized the publicly available~\cite{wang2022transformer,transpath} ViT-Small network pretrained using MoCo v3~\cite{mocov3} on all the slides from TCGA~\cite{weinstein2013cancer} and PAIP~\cite{paip}. 
In Table~\ref{table:result:universal_models}, we showed that our method robustly boosted the performance on both TCGA (the same domain as the foundation model trained on) and BRIGHT (a different domain). 
The improvement is more prominent in BRIGHT, which further confirmed that Prompt-MIL aligns the feature extractor to be more task-specific.

\begin{table}[h]
\caption{Performance with a different number of prompt tokens. For two different WSI classification tasks, one token was enough to boost the performance of the conventional MIL schemes.}
\label{table:result:abalation}
\begin{center}
\setlength{\tabcolsep}{1.6mm}{

\begin{tabular}{ccccc}
\toprule
\multicolumn{1}{c}{Dataset} 
    & \multicolumn{2}{c}{TCGA-BRCA}        
        & \multicolumn{2}{c}{BRIGHT}           \\
\#prompt tokens $k$             
    & \multicolumn{1}{l}{Accuracy} & AUROC 
        & \multicolumn{1}{l}{Accuracy} & AUROC \\ 
\midrule
$k=1$                           
    & $\bm{93.47}$  & $\bm{96.89}$ 
        & $\bm{64.58}$  & $\bm{81.31}$ \\
$k=2$                       
    & 93.13   & $\bm{96.93}$ 
        & 60.41 & 79.74 \\
$k=3$                           
    &  $\bm{93.47}$  &  $96.86$     
        &  59.17   &  76.75     \\ 
\bottomrule
\end{tabular}
}
\end{center}
\end{table}

\resultsection{Ablation study:}
An ablation was performed to study the effect of the number of trainable prompt tokens on downstream tasks. 
Table~\ref{table:result:abalation} shows the accuracy and AUROC of our Prompt-MIL model with 1, 2 and 3 trainable prompt tokens ($k=1, 2, 3$) on the TCGA-BRCA and the BRIGHT datasets.
On the TCGA-BRCA dataset, our Prompt-MIL model with 1 to 3 prompt tokens reported similar performance.
On the BRIGHT dataset, the performance of our model dropped with the increased number of prompt tokens. 
Empirically, this ablation study shows that for classification tasks, one prompt token is sufficient to boost the performance of conventional MIL methods.

%% file: 4_sum.tex
\section{Conclusion}
In this work, we introduced a new framework, Prompt-MIL, which combines the use of Multiple Instance Learning (MIL) with prompts to improve the performance of WSI classification.  Prompt-MIL adopts a prompt tuning mechanism rather than a conventional full fine-tuning of the entire feature representation. In such a scheme, only a small fraction of parameters calibrates the pretrained representations to encode task-specific information, so the entire training can be performed in an end-to-end manner. 
We applied our proposed method to three publicly available datasets.
Extensive experiments demonstrated the superiority of Prompt-MIL over the conventional MIL as well as the conventional fully fine-tuning methods.
Moreover, by fine-tuning much fewer parameters compared to fully fine-tuning, our method is GPU memory efficient and fast.
Our proposed approach also showed promising potentials in transferring foundation models. 
We will further explore the task-specific features that are captured by our prompt toward explainability of these models.